**A Computationally Efficient Multi-Section Whole Slide Image Analysis for Capturing Prostate Cancer Multifocality in Biochemical Recurrence Prediction**

Yesung Cho[1]*, Dongmyung Shin[1,2]*, Sujeong Hong[1], Jooyeon Lee[6], Seongmin Park[2], Geongyu Lee[1], Won Young Park[3], Bongjin Park[4], Jongbae Park[2,5+], Hong Koo Ha[6+]

*Equal contribution. +Corresponding authors.*

[1]OmixAI Co. Ltd., Seoul, Republic of Korea. [2]Oncocross Co. Ltd., Seoul, Republic of Korea. [3]Seegene Medical Foundation, Busan, Republic of Korea. [4]Department of Neurosurgery, Kyung Hee University College of Medicine, Seoul, Republic of Korea. [5]Department of Basic Biomedical Sciences, Kyung Hee University, Seoul, Republic of Korea. [6]Department of Urology, School of Medicine, Pusan National University, Republic of Korea.

**Abstract —** Prostate cancer is among the most frequently diagnosed malignancies in men worldwide. However, precise prediction of biochemical recurrence (BCR) after radical prostatectomy remains challenging. Here, we propose a computationally efficient multi-section whole-slide image analysis framework designed to capture the whole-gland tumor landscape. We curated a large-scale dataset comprising 23,451 slides from 789 patients and systematically optimized patch and slide sub-sampling to substantially reduce the computational cost of training and inference. Under this efficient configuration, the framework achieved strong predictive performance for 1- and 2-year BCR, with discrimination and clinical net benefit comparable to or exceeding those of established clinical benchmarks. In multivariable Cox proportional hazards analysis, the AI-derived risk score remained an independent prognostic factor among the covariates examined, and its integration with clinical variables further improved prognostic performance. External validation in an independent cohort provided preliminary evidence of transferability. Finally, attention-dispersion analysis showed that the model drew prognostic information from multiple sections, consistent with the multifocal nature of prostate cancer. Collectively, these findings support the potential of multi-section slide analysis as a scalable approach for postoperative BCR risk assessment in prostate cancer.

## 1. Introduction

Prostate cancer remains one of the most prevalent cancers in men worldwide, with millions of new diagnoses annually [1,2]. Primary treatments often involve radical prostatectomy (RP) or radiotherapy combined with androgen deprivation therapy, depending on cancer stage and patient health [3–6]. However, even after treatment, a substantial proportion of patients experience biochemical recurrence (BCR), characterized by elevated serum levels of prostate-specific antigen (PSA) [7,8]. Therefore, there has been an increasing demand to predict the BCR outcomes of patients to guide the patient's treatment strategy, including the scheduling of surveillance intervals and the administration of adjuvant therapies [9–11]. Predicting early BCR is particularly important, as short-term recurrence strongly correlates with accelerated disease progression, elevated metastatic risk, and the eventual development of castration-resistant prostate cancer [12].

Previous studies have proposed many promising methods to stratify patients' risks of BCR based on clinical and/or pathological features [11,13–15]. One of the well-known methods is Han Tables [15], which calculates the probability of BCR for up to 10 years using the PSA level, Gleason score, and pathological cancer stage. Furthermore, CAPRA-S [13] integrates additional information, such as surgical margin status and lymphatic invasion, for patient stratification [13,16,17]. Similarly, Bergero et al. [17] developed a new strategy to accurately predict future risks of BCR using a combination of 16 clinical and pathological factors.

More recently, with the emergence of artificial intelligence (AI), computational pathology has been applied to BCR prediction [18]. For example, Cao et al. [19] developed a deep learning framework to predict the that analyzes whole slide images (WSIs) of prostate tissues to identify prognostic features related to BCR.

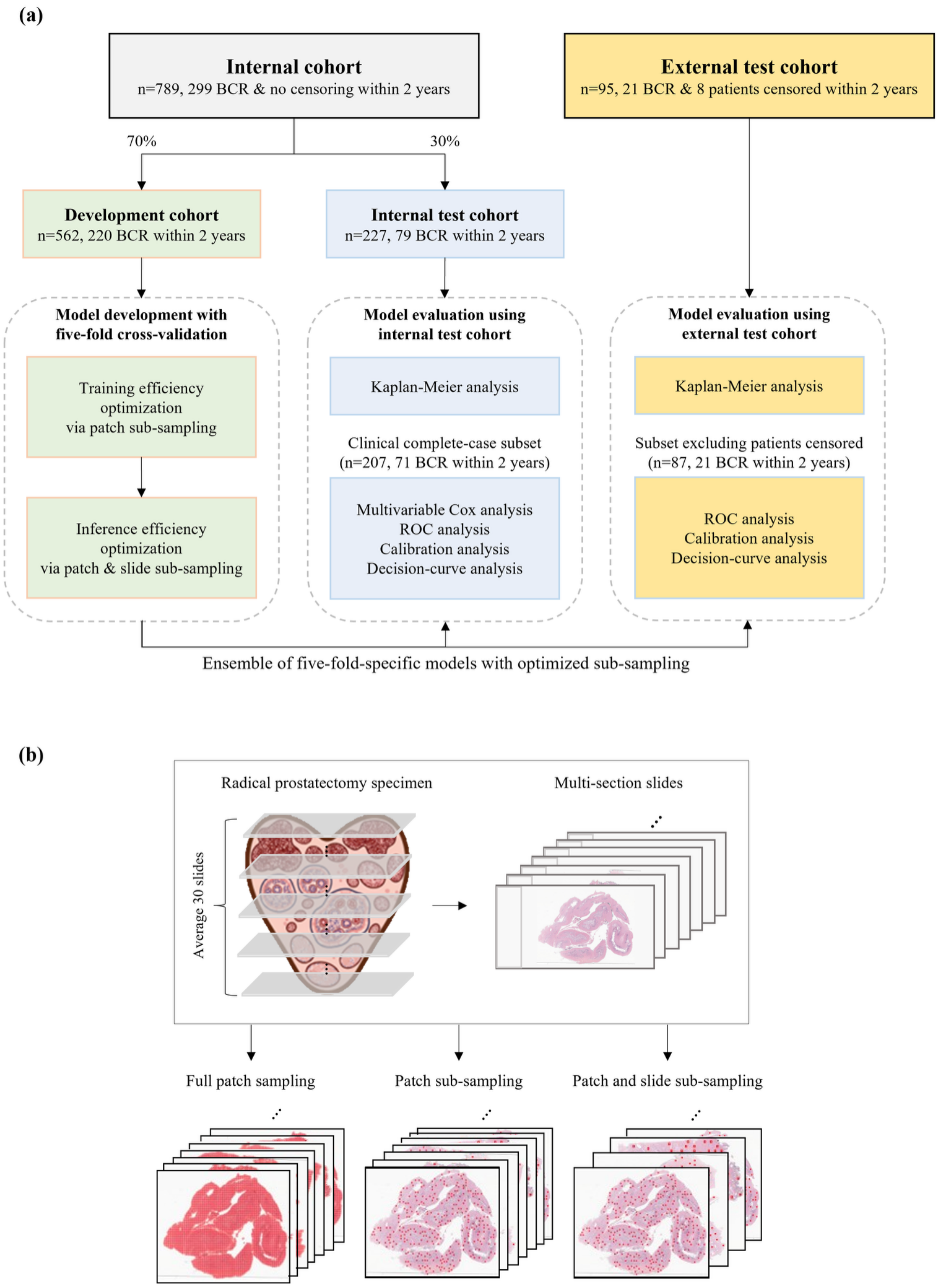

Fig. 1. An overview of study design, data pipeline, and multi-section sampling strategies. (a) Study design and data analysis pipeline. An internal cohort (789 patients, 299 with BCR; 23,451 slides) was split 7:3 into a development cohort (562 patients, 220 BCR) and an internal test set (227 patients, 79 BCR). Within the development cohort, five-fold cross-validation was used to optimize training-data efficiency (patch sub-sampling ratio) and inference efficiency (patch and slide sub-sampling), yielding the ensemble model with optimized sub-sampling ratios applied to all subsequent evaluations. In the internal test cohort, Kaplan-Meier analysis used 227 patients, whereas the multivariable Cox and other analyses (ROC, calibration, and decision curve analysis) used clinical complete-case subset (20 excluded for missing clinical data). For external validation, the CHIMERA cohort (95 patients, 21 BCR) was reduced to 87 patients after excluding 8 patients censored within the 2-year window and evaluated by ROC, calibration, and DCA analyses. (b) Sampling strategies for multi-section slides. Each radical prostatectomy specimen yields multiple whole-slide sections (~30 per patient on average). Full patch sampling processes all extracted patches (high computational cost); patch sub-sampling retains an informative subset of patches; and patch-and-slide sub-sampling additionally selects a subset of sections, further reducing computational cost.

However, all previous works are restricted to single-region analysis, relying on an individual TMA core [21–23], biopsy [19,24], or WSI from RP [20,25] for each patient, which creates a spatially fragmented view that neglects the multifocality of prostate cancer [8,26]. Unlike other malignancies, prostate cancer manifests as distinct tumor foci with heterogeneous characteristics distributed throughout the prostate gland [8,9]. Therefore, accurate BCR prediction requires computational approaches capable of integrating pathological features from spatially distributed tumor regions. To overcome these limitations, we propose a computationally efficient multi-section WSI analysis framework that captures whole-prostate multifocality for patient-level BCR prediction. This framework is designed to extract and integrate prognostic pathological features across the entire prostate gland, while substantially reducing computational demands for AI training and testing through patch and slide sub-sampling strategies. To develop this, we have curated 23,451 WSIs from 789 patients' tissues after RP.

Our contributions are as follows:

1) Development of a computationally efficient multi-section WSI analysis framework that captures whole-prostate multifocality using a large-scale, curated dataset.
2) Optimization of computational cost via patch and slide sub-sampling strategies, significantly accelerating both training and inference efficiency without compromising predictive performance.
3) Comprehensive prognostic validation and interpretability of the AI-derived risk score, demonstrating superior and complementary prognostic value over established clinical benchmarks while showing that the model attention is distributed across multiple whole-gland sections.

## 2. Materials and Methods

### *2.1. Patient Population and Characteristics*

Multi-section WSIs derived from RP specimens were retrospectively collected from an anonymized tertiary academic medical center between February 2010 and May 2025. Patients underwent RP between February 2010 and March 2021 and were followed through the study cutoff date of 10 May 2025. Given that the most recent surgery was performed in 2021, all patients had at least approximately four years of follow-up. Accordingly, no patient was censored within 2 years after RP. BCR was defined as two consecutive postoperative PSA measurements $\geq 0.2$ ng/mL. The cohort included 789 patients (mean age: $66.7 \pm 6.0$ years), with 299 (37.9 %) experiencing BCR within 2 years (mean time to BCR: $6.98 \pm 6.6$ months) and 490 remaining BCR-free until then. The dataset consisted of a total of 23,451 slides, with an average of 29.7 slides per patient. For model training and evaluation, the cohort was partitioned into a development cohort (n = 562) and a test cohort (n = 227) at a 7:3 ratio (see Fig. 1a).

In addition to BCR outcomes, several clinicopathological variables were collected for each patient, including pre-operative PSA, pathological stage, Gleason score, surgical margin status, lymphatic invasion, tumor percentage and number of positive lymph nodes. Detailed information for all the variables and their statistics is provided in Supplementary Table S1. The pathological parameters were assessed by a board-certified pathologist in the medical center.

For external validation, we utilized the publicly available CHIMERA dataset [27]. The initial cohort comprised 95 patients, where 21 patients (22.1%) experienced BCR within 2 years and others remained BCR-free until then (see Fig. 1a). In CHIMERA, BCR is defined as a postoperative PSA $\geq 0.1$ ng/mL, in contrast to the PSA $\geq 0.2$ ng/mL threshold used for our internal cohort.

### *2.2. AI Model Training using Density-Based Patch Sampling*

For AI development, we adopted an attention-based deep multiple instance learning (MIL) framework [28], using non-overlapping WSI patches with patient-level labels (i.e., BCR outcome). The model was trained on the 2-year BCR label and yields a single continuous per-patient risk score; the reported 1- and 2-year performances were obtained by evaluating this same score against the respective ground-truth definitions. Typically, standard WSI datasets represent a patient with a few WSIs, allowing the use of all patches for AI training without significant computational burden. However, the dataset used in this study consists of multi-section WSIs covering the entire prostate gland. Therefore, utilizing every patch for training is computationally prohibitive, as the average slide count per patient is 29.7 (see 'Full patch sampling' in Fig. 1b). To address this tremendous computational cost, we employed a patch sub-sampling method based on tissue density [29]. This sampling strategy prioritizes tissue-dense regions, thereby increasing the likelihood of selecting morphologically informative areas, while substantially reducing the patch count compared to the full patch sampling strategy (see 'Patch sub-sampling' in Fig. 1b).

Let $x_i \in R^2$ denote the center coordinate of the $i^{th}$ patch, and $N$ denote the total number of patches in a WSI. Then we estimate the sampling probability of the $i^{th}$ patch ($p_i$) as follows:

$$p_i = \frac{\hat{f}(x_i)}{\sum_{j=1}^{N} \hat{f}(x_j)},\ where\ \hat{f}(x) = \frac{1}{Nh^2} \sum_{k=1}^{N} K(\frac{x - x_k}{h}) \quad (1)$$

Here, $K(\cdot)$ is a Gaussian kernel function, and $h$ denotes the bandwidth set to the patch size (= 512). The term $\hat{f}(x)$ represents the estimated patch density. By normalizing these density values, we derive the probability distribution of individual patches. Based on these probabilities, we sampled a fixed fraction of patches per slide.

### *2.3. Optimization of Training Efficiency with Varying Patch Sub-Sampling Ratios*

First, we assessed the AI model's predictive performance across different training patch sub-sampling ratios (10%, 20%, and 30%; see 'Training efficiency optimization' in Fig. 1a). For each ratio, predictive performance was evaluated on a five-fold cross-validation: within each fold, predictions were generated for the held-out validation split, and the out-of-fold predictions from all five folds were pooled to yield one prediction per development-cohort patient (n=562). Therefore, no test cohort data was used at this stage. To ensure a fair comparison of predictive performance across different training patch sub-sampling ratios, inference was consistently performed using full patch sampling. Performance was quantified using the area under the receiver operating characteristic curve (AUC) with 95% confidence intervals (CI) calculated by the bootstrap method with 2,000 iterations. To identify the most data-efficient configuration, the smallest ratio (10%) served as the reference, and DeLong's test [30] assessed whether the higher ratios (20% and 30%) yielded significantly different pooled AUCs from this reference.

### *2.4. Optimization of Inference Efficiency with Varying Patch and Slide Sub-Sampling Ratios*

To optimize inference efficiency of AI, we evaluated the trade-off between inference time and predictive accuracy for 2-year BCR across varying patch and slide sub-sampling ratios (see 'Inference efficiency optimization' in Fig. 1a). Same as in the training-efficiency analysis, predictive performance was quantified on the pooled out-of-fold validation predictions in the development cohort (n=562). First, we systematically varied the inference patch sub-sampling ratio from 0.05% to 30% while recording both predictive performance (in AUC) and average inference time per patient (in minutes). The same density-based sampling method [29] employed during training was utilized for inference. Second, to quantify the impact of spatial coverage of slides on predictive accuracy, we incrementally added the number of inference slides, while keeping the inference patch sub-sampling ratio fixed (see 'Patch and slide sub-sampling' in Fig. 1b). We compared two slide-selection strategies: random sampling, where slides were selected arbitrarily, and uniform sampling, which selected the slides in equidistant intervals to maximize anatomical coverage. For both strategies, we produced performance curves, plotting AUC and average inference time per patient against the number of input slides. For both experiments, sub-sampling was repeated (three times per patch ratio and five times per slide count), and AUCs are reported as the mean with the 95% confidence interval across repetitions (shaded bands, Fig. 2c,d).

### *2.5. Construction of Five-Fold Ensemble Model and Locking*

Following the efficiency optimization, the training and inference sampling ratios that achieved the best computational efficiency while maintaining predictive performance were selected. The five-fold-specific models obtained under these optimized sampling settings were then retained and combined into a single locked ensemble (see Fig. 1a). For each patient in the internal (Section 2.6 and 2.7) and external (Section 2.8) test cohorts, predictions from the five models were averaged to generate the final AI risk score. The selected sampling ratios and all model parameters were fixed before test-cohort evaluation, and no further training, model selection, or tuning was performed using the test data.

### *2.6. Kaplan-Meier and Multivariable Cox Analyses*

Kaplan–Meier recurrence-free survival analysis with log-rank testing was performed on the internal test cohort (n = 227; see Fig. 1a), with patients partitioned into low- and high-risk groups by the median AI risk score derived from the development cohort (threshold=0.458).

To evaluate the independent prognostic value of the AI prediction for 2-year BCR, we fitted a multivariable Cox proportional hazards regression on the internal test complete-case subset (n = 207), incorporating the AI risk score alongside age, pre-operative PSA, pathological T stage, Gleason score, surgical margin status, lymphatic invasion, tumor percentage, and positive lymph node count. Hazard ratios (HR) with 95% CI and Wald-test [32] significance were derived for each variable. The proportional-hazards assumption for this model was assessed using Schoenfeld residuals, with no significant violation observed for any covariate (all $p > 0.05$), including the AI risk score ($p = 0.81$).

Also, to compare the discriminative performance of the AI, clinical, and combined predictors, we additionally constructed three Cox models: (1) an AI-only model containing the AI risk score as the single covariate; (2) a clinical-only multivariable model containing the eight clinicopathological variables listed above; and (3) a combined model containing both. To ensure a consistent comparison across models, all three Cox models were fitted on the development cohort with complete clinicopathological data (n = 518) and then evaluated in the held-out internal test subset (n = 207) using Harrell's concordance index (C-index).

### *2.7. ROC, Calibration, and DCA: Comparison with Other Clinical Models*

We compared the predictive performance of the locked ensemble model with the CAPRA-S [13] and XGBoost model [17] for both 1-year and 2-year BCR prediction through ROC analysis (bootstrap with 2,000 iterations; DeLong's test for statistical significance), calibration plot (with Brier and Brier skill scores), and decision-curve analysis (DCA) on the clinical complete-case subset (n = 207). The locked ensemble model was used as the reference for pairwise comparisons.

For XGBoost, we performed a partial-feature reimplementation of Bergero et al. [17] due to differences in data availability. We could utilize 9 out of 16 clinicopathological features used in the original study (see Supplementary Table S2). To avoid optimistic tuning, we adopted the original study's hyperparameters without modification (max depth = 2, eta = 0.2, rounds = 23). Mirroring the ensemble model, five XGBoost models were trained on the same five cross-validation splits of the development cohort, and their predicted probabilities were averaged for the test cohort evaluation. Categorical variables (e.g., pathological T stage) were one-hot encoded jointly across the development and test cohorts to ensure identical feature columns.

CAPRA-S was calculated according to Cooperberg et al. [13] using six postoperative variables: preoperative PSA, pathological Gleason grade, surgical margin status, extracapsular extension, seminal vesicle invasion, and lymph node involvement. All six required variables were available in the clinical complete-case subset. The resulting scores, ranging from 0 to 12, were linearly rescaled to the range of 0 to 1 by dividing each score by 12 and were used as normalized risk scores for model comparison.

### *2.8. External Validation of AI Model*

The generalizability of the locked ensemble model was assessed on the external CHIMERA (n = 95) dataset using the same 1% patch sub-sampling ratio as in the internal cohort (see Fig. 1a). It is worth noting that the CHIMERA cohort contained an average of approximately 1.9 WSIs per patient, and the fixed 19-slide subsampling strategy was not applicable. Therefore, all available sections for each patient were used as model inputs (see Discussion).

Kaplan–Meier recurrence-free survival analysis was performed in the full external cohort, with censoring handled according to the observed follow-up time. Patients were stratified into low- and high-risk groups using the median AI risk score prespecified from the internal development cohort.

For the 2-year discrimination, calibration, and decision curve analyses, eight patients were censored within 2 years and excluded, resulting in an evaluable cohort of 87 patients, including 21 patients with BCR and 66 without BCR (Fig. 1a). For the 1-year analysis, 4 patients were censored, resulting in a total of 91 patients with 17 BCR occurrences. All statistical procedures were consistent with those used for the internal test cohort.

### *2.9. Assessment of Whole-Gland Attention Dispersion and Pathological Correlates*

To examine whether the model's prognostic evidence was distributed across the whole gland or concentrated in a few sections, we quantified how the model's recurrence-associated attention was dispersed across each patient's multi-section WSIs. For each patient in the internal test cohort, patch-level attention weights were extracted from the BCR output branch and normalized to sum to one across all patches included in the inference. We then retained the top 10% most-attended patches and computed $\boldsymbol{p} = \{p_1, p_2, \ldots, p_S\}$ which denotes the proportion of these patches across the patient's total $S$ slides ($\sum_{s=1}^{S} p_s = 1$).

Then, we characterized the dispersion using the following metric:

$$N_{eff}(\boldsymbol{p}) = e^{H(\boldsymbol{p})}, \;\; H(\boldsymbol{p}) = -\sum_{s=1}^{S} p_s \ln p_s \qquad (2)$$

where $H(\boldsymbol{p})$ is the Shannon entropy of the slide-level distribution, and $N_{\text{eff}}$ represents the effective number of slides contributing to the model prediction. $N_{\text{eff}} = 1$ indicates that the selected high-attention patches are concentrated on a single slide, whereas $N_{\text{eff}} = S$ indicates that they are uniformly distributed across all available slides. To visualize the extent of whole-gland dispersion, $N_{\text{eff}}$ was plotted against the total number of slides, $S$, with each point representing one patient. This analysis was performed in 226 patients from the internal test cohort (n = 227) after excluding one patient for whom only a single slide was available.

To qualitatively assess whether spatially dispersed prognostic attention corresponded to histologically distinct tumor foci, we performed case-level pathological reviews. Two patients with the highest $N_{\text{eff}}$ values were selected as representative cases exhibiting broadly distributed attention across whole-gland sections. For each patient, slides were ranked according to their total attention mass, calculated as sum of the patch-level attention weights within each slide. The six highest-ranked slides were retained, and the patch with the highest attention weight from each slide was extracted for pathological review.

A board-certified pathologist reviewed the selected regions according to a predefined assessment framework. For each region, the pathologist first determined the tissue type, distinguishing carcinoma from benign glandular tissue, stroma, and artifacts. Tumor-containing regions were further evaluated for Gleason pattern and associated architectural morphology (e.g., fused or cribriform morphology), nuclear characteristics such as nucleolar prominence, and other notable histological findings.

### *2.10. Implementation Details*

Internal H&E-stained WSIs were scanned on an Aperio GT 450 DX system (Leica Biosystems, IL, USA). Tissue regions were segmented using Otsu thresholding [33], and non-overlapping 512 × 512 patches were extracted at 40× magnification, without staining normalization. Density-based patch subsampling was first performed [29], followed by patch-level feature extraction using the UNI foundation model [34]. These features served as input to the attention-based MIL framework [28] with two attention branches corresponding to the BCR and non-BCR outputs, and the final prediction was the class with the higher probability. The model was trained for 100 epochs with a cross-entropy loss and an additional L2 regularization term. Optimization used the Adam optimizer [35] (learning rate $5\times10^{-6}$, weight decay $5\times10^{-7}$) with gradient accumulation over 16 single-patient iterations (i.e., an effective batch of 16 patients per update); the dropout rate was 0.25, and the random seed was fixed at 7 for reproducibility. The training and inference codes are provided in the code repository. Experiments were implemented in Python 3.11 and PyTorch 2.0.1 (CUDA 11.7) and run on a workstation with an Intel Xeon E5-2698 v4 CPU and four NVIDIA Tesla V100 GPUs (32 GB VRAM).

## 3. Results

### *3.1. Optimization of Training Efficiency with Varying Patch Sub-Sampling Ratios*

Figure 2a and 2b present the model's five-fold cross-validation results for 1- and 2-year BCR across varying training patch sub-sampling ratios (10%, 20%, and 30%) in the development cohort (n=562). Notably, the model maintained consistent performance regardless of the sampling ratio, yielding AUCs of 0.85 and 0.82 for the 1-year and 2-year endpoints, respectively. Pairwise statistical analyses revealed no significant differences in predictive accuracy between the 10% sub-sampling model and those trained with higher ratios ($p > 0.05$ for all comparisons). Because increasing the sampling ratio to 20% or 30% provided no additional performance gains, the 10% ratio was selected as the best training configuration, which maintains predictive accuracy while minimizing training cost.

### *3.2. Optimization of Inference Efficiency with Varying Patch and Slide Sub-Sampling Ratios*

Figure 2c illustrates the trade-off between 2-year BCR predictive performance (in AUC) and inference time

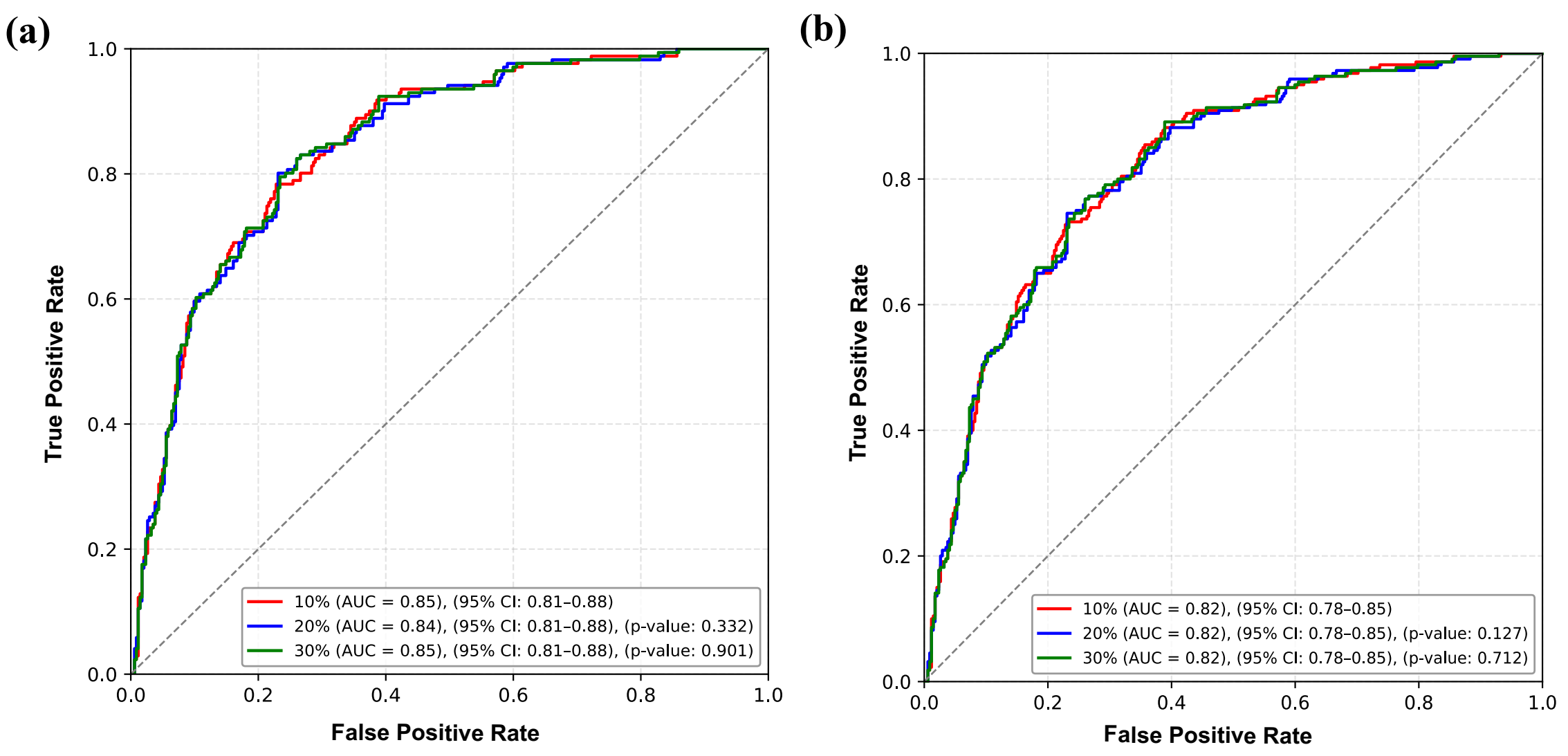

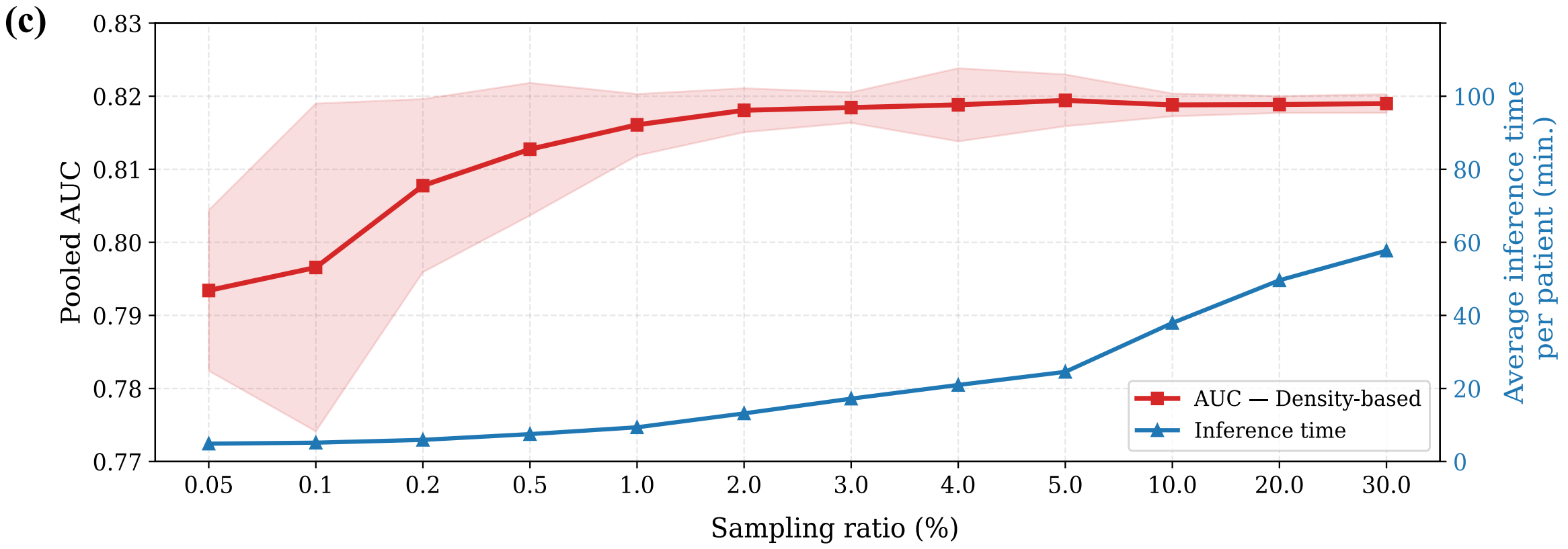

| | | | | | | | | | | | | |
|---|---|---|---|---|---|---|---|---|---|---|---|---|
| **AUC (density)** | 0.79 | 0.80 | 0.81 | 0.81 | 0.82 | 0.82 | 0.82 | 0.82 | 0.82 | 0.82 | 0.82 | 0.82 |
| **Time (min.)** | 4.9 | 5.2 | 5.9 | 7.5 | 9.4 | 13.2 | 17.2 | 21.0 | 24.5 | 37.9 | 49.6 | 57.7 |

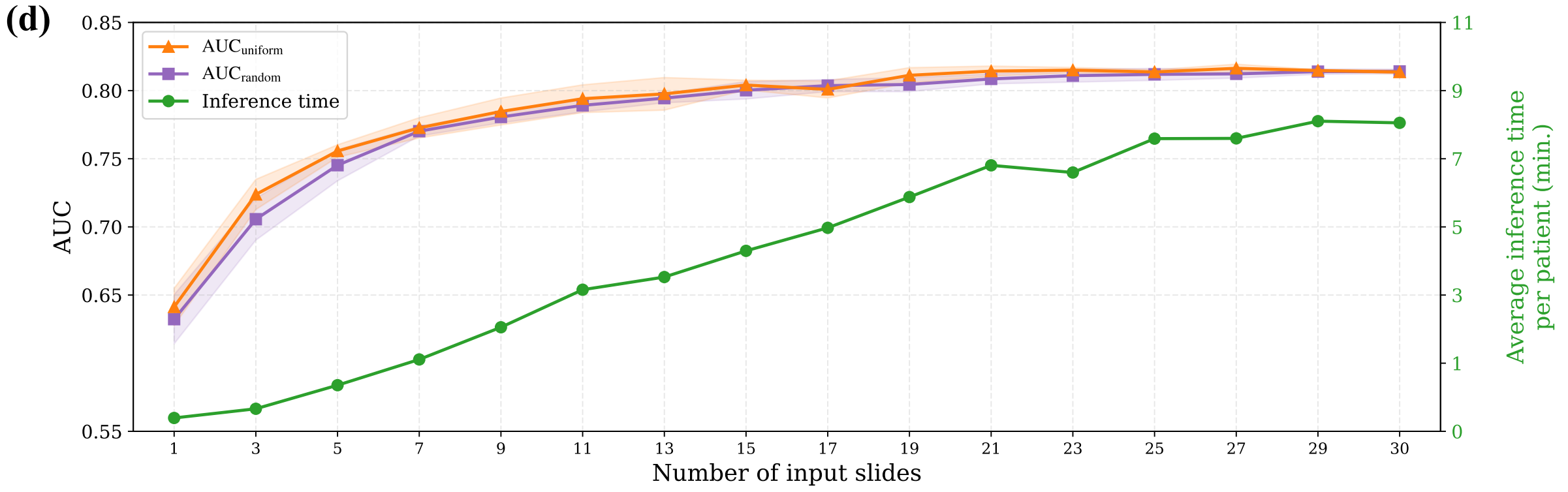

| | | | | | | | | | | | | | | | | |
|---|---|---|---|---|---|---|---|---|---|---|---|---|---|---|---|---|
| $AUC_{uniform}$ | 0.64 | 0.72 | 0.76 | 0.77 | 0.78 | 0.79 | 0.80 | 0.80 | 0.80 | 0.81 | 0.81 | 0.81 | 0.81 | 0.82 | 0.81 | 0.81 |
| $AUC_{random}$ | 0.63 | 0.71 | 0.75 | 0.77 | 0.78 | 0.79 | 0.79 | 0.80 | 0.80 | 0.80 | 0.81 | 0.81 | 0.81 | 0.81 | 0.81 | 0.81 |
| Time (min. ) | 0.4 | 0.6 | 1.2 | 1.9 | 2.8 | 3.8 | 4.2 | 4.9 | 5.5 | 6.3 | 7.2 | 7.0 | 7.9 | 7.9 | 8.3 | 8.3 |

Fig. 2. Results of computational efficiency optimization. (a, b) ROC curves for 10%, 20%, and 30% training patch sub-sampling ratios, for 1-year (a) and 2-year (b) BCR, obtained by pooling held-out validation predictions across the five cross-validation folds in the development cohort (n = 562). Discrimination was comparable across ratios (AUC 0.84–0.85 at 1 year; 0.82 at 2 years; DeLong's test vs. the 10% reference, $p > 0.05$). (c) Two-year BCR discrimination (AUC) and average inference time per patient versus the inference patch sub-sampling ratio (0.05–30%); lines show the mean and shaded bands the 95% CI across three repeated sub-samplings. Performance plateaued nearly 1% (AUC 0.82, 9.4 min). (d) AUC and inference time versus the number of input slides for uniform and random slide sampling; lines show the mean and shaded bands the 95% CI across five repeated sub-samplings. Uniform sampling reached plateaued accuracy at 19 slides (AUC 0.81, 6.3 min/patient).

(in minutes) across varying inference patch sampling ratios. The AUC improved from 0.80 to 0.82 as the sampling ratio increased from 0.05% to 30%. Interestingly, the model retained substantial predictive power even at a minimal sampling ratio of 0.05% (AUC = 0.80), suggesting that most prognostic features are driven by a tiny fraction of highly informative patches. Inference time scaled with sampling ratios, requiring 9.4 and 13.2 minutes at 1% and 2% ratios, respectively. These results indicate that exhaustive patch sampling of multi-section slides during inference is unnecessary, and, consequently, a sampling of 1% is optimal to achieve peak performance (AUC = 0.82) with substantially reduced computational overhead.

Building upon our patch-level inference efficiency optimization, we investigated the effect of slide-level coverage on both predictive performance and inference speed. As depicted in Figure 2d, both uniform and random slide sampling strategies exhibited a steady increase in AUC, peaking at 19 and 21 slides, respectively, before plateauing. Notably, the uniform sampling strategy achieved an AUC of 0.81 using only 19 slides, which consequently reduced the average inference time per patient to 6.3 minutes. These findings demonstrate that a representative subset of 19 evenly sampled slides adequately captures the prognostic information of the full multi-section slide set, substantially alleviating the computational burden without compromising predictive accuracy. Taken together, a patch sub-sampling ratio of 1% combined with 19 uniformly sampled slides was selected as the optimal inference configuration, which was adopted for all subsequent analyses.

### *3.3. Kaplan-Meier and Multivariable Cox Analyses*

Figure 3a presents the multivariable Cox proportional hazards analysis (forest plot) evaluating the prognostic value of the AI-derived risk score alongside clinicopathological variables for 2-year BCR. The analysis identified the AI risk score as a statistically significant independent predictor of BCR, with the largest adjusted hazard ratio among the covariates examined (HR 3.27; 95% CI 1.67–6.41; $p < 0.001$). Among the clinicopathological variables, pathological Gleason score (HR: 1.29, p=0.011) and lymphatic invasion (HR: 1.73, p=0.044) remained significant. Here, pathological T stage 2b and T stage 4 were excluded from the analysis since they were not represented in the internal test cohort (See Supplementary Table S1).

The risk stratification capability of the AI score was further validated using Kaplan–Meier survival analysis (Fig. 3b). The low-risk group (n = 134) maintained a high recurrence-free survival rate of approximately 88% at 24 months. In contrast, the high-risk group (n = 93) exhibited a substantially lower survival rate of roughly 37% at the same time point (log-rank $p < 3.2 \times 10^{-17}$). This pronounced prognostic separation confirms the clinical utility of the AI risk score as a robust binary stratification tool for identifying patients at an elevated risk of BCR.

When we compared clinical-only, AI-only, and combined clinical-plus-AI models, the combined model achieved the highest discriminative performance (C-index, 0.79), followed closely by the AI-only model (C-index, 0.78), while the clinical-only model showed a lower C-index of 0.76. These results possibly imply that the AI risk score captures most of the prognostic signal and provides strong prognostic discrimination while remaining complementary to standard clinicopathological variables.

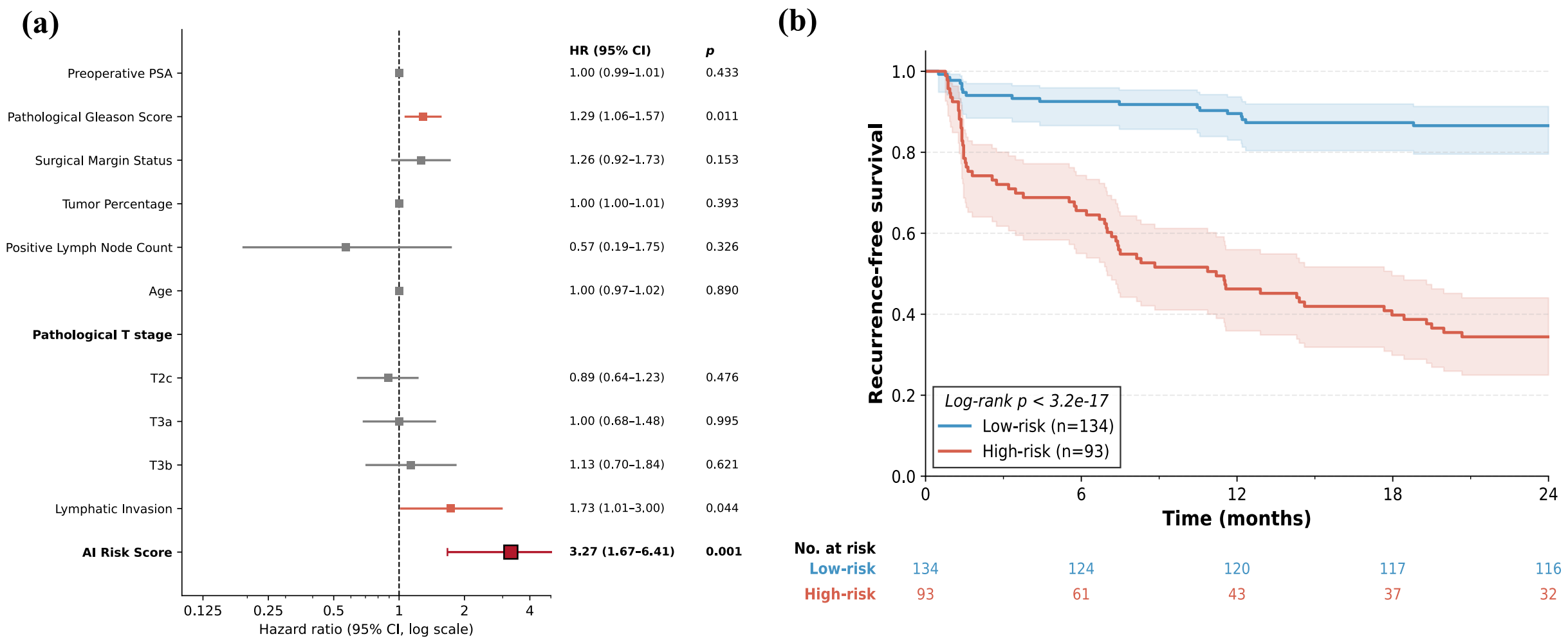

Fig. 3. Multivariable Cox and KM analyses of the AI risk score on the internal test cohort. (a) Forest plot of the multivariable Cox proportional hazards analysis for 2-year BCR, showing hazard ratios (95% CI) for the AI risk score and clinicopathological variables (complete-case subset, n = 207). The AI risk score was the strongest independent predictor (HR 3.27, 95% CI 1.67–6.41). The dashed line marks HR = 1. (b) Kaplan–Meier recurrence-free survival stratified by the AI risk score (n = 227), where patients were dichotomized into low-risk (n = 134) and high-risk (n = 93) groups at the median training-cohort score (threshold = 0.458), with shaded regions denoting 95% confidence intervals.

### *3.4. ROC, Calibration, and DCA: Comparison with Other Clinical Models*

Predictive performance for 1-year and 2-year BCR was comparatively analyzed between our AI model and existing clinical models (Fig. 4a,b). For the 1-year endpoint, the AI model achieved an AUC of 0.84, compared with 0.77 for CAPRA-S and 0.77 for XGBoost reimplementation; both differences were statistically significant (DeLong's test, p=0.02). For the 2-year endpoint, the AI model attained an AUC of 0.84, versus 0.79 for CAPRA-S and 0.78 for XGBoost; although numerically higher, neither difference reached statistical significance (XGBoost, p = 0.06; CAPRA-S, p = 0.08).

Beyond discrimination, we compared the calibration and clinical utility of the models for predicting 2-year BCR (Fig. 4c,d). Consistent with the calibration plot, the AI model demonstrated the lowest Brier score of 0.158, and the highest Brier skill score of 0.30, compared with CAPRA-S (Brier score: 0.171; Brier skill score: 0.24) and XGBoost (Brier score: 0.176; Brier skill score: 0.22). On the calibration curves, the three models were broadly comparable, with largely overlapping confidence bands. The AI model somewhat over-estimated risk in the lower predicted-probability range, with its calibration curve progressively approaching the diagonal (perfect calibration) at intermediate and higher predicted probabilities. Decision curve analysis further showed that the AI model provided a greater net benefit than CAPRA-S, XGBoost, and the treat-all, and treat-none strategies across threshold probabilities of **20%–80%**.

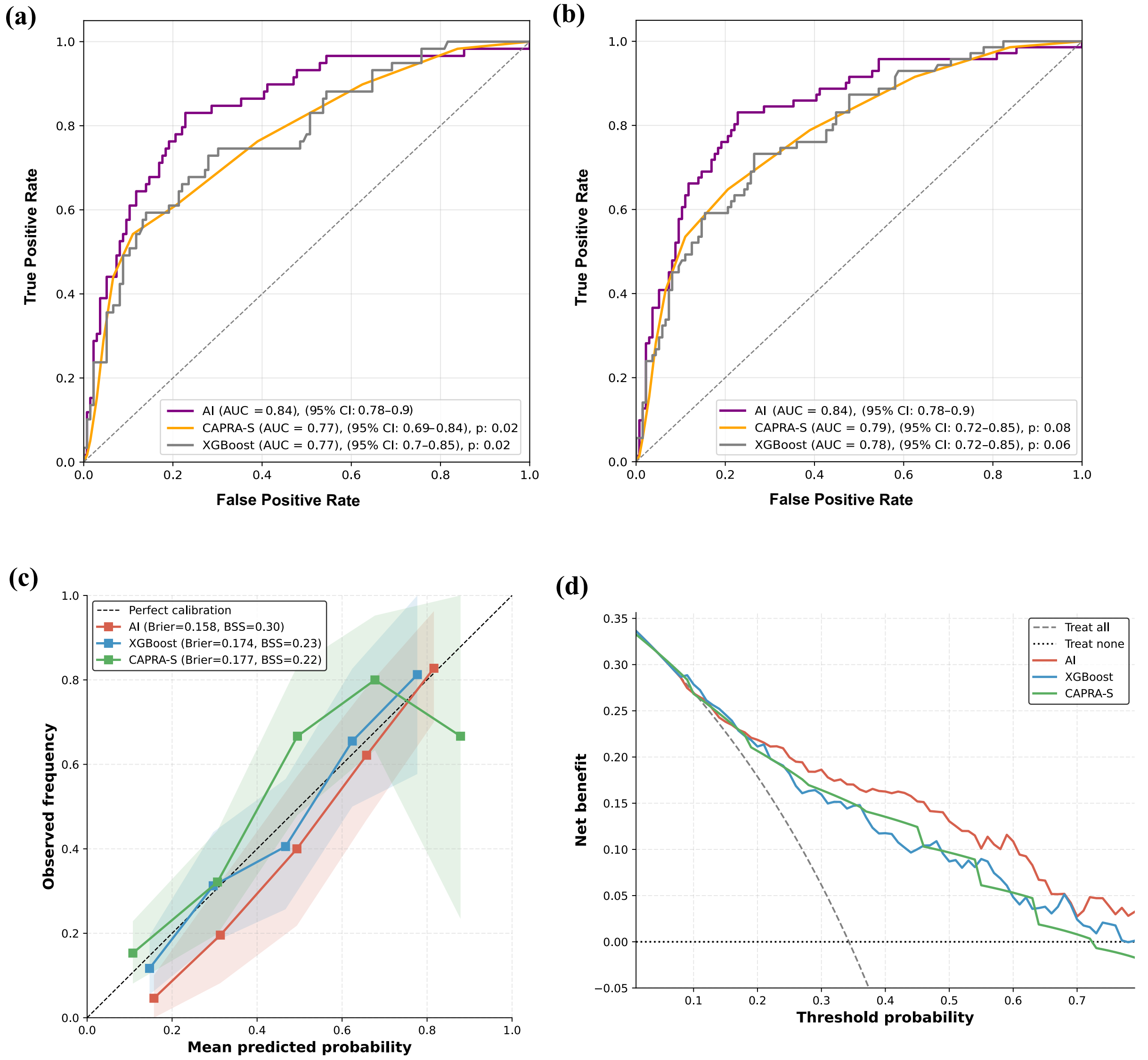

Fig. 4. Comparison of the AI model with established clinical benchmarks (internal test cohort, n = 207). (a, b) ROC curves for the AI score, CAPRA-S, and XGBoost [17] for 1-year (a) and 2-year (b) BCR prediction. The AI model achieved the highest AUC at both horizons. At 1 year, it significantly outperformed CAPRA-S (AUC 0.77) and XGBoost (AUC 0.77; DeLong's test, p = 0.02 for both); at 2 years, the AI model showed a numerically higher AUC than XGBoost (0.78, p = 0.06) and CAPRA-S (0.79, p = 0.08), although neither difference reached significance. (c) Calibration plot for 2-year BCR (dashed diagonal = perfect calibration). The AI model achieved the best overall probabilistic accuracy (Brier 0.158, BSS 0.30) versus CAPRA-S (0.171, 0.24) and XGBoost (0.176, 0.22). (d) Decision curve analysis for 2-year BCR; the AI model provided the highest net benefit across threshold probabilities of 20%–80%, exceeding both benchmarks, treat-all, and treat-none reference strategies.

### *3.5. External Validation of AI Model*

Figure 5a summarizes the external validation results of Kaplan–Meier analysis on the external test cohort (n = 95; see Fig. 1a). The high-risk group showed substantially poorer recurrence-free survival than the low-risk group, with estimated 2-year recurrence-free survival rates of approximately 37% and 88%, respectively. The difference was statistically significant (log-rank $p < 0.001$), and the high-risk group had a higher risk of recurrence in the Cox model with follow-up administratively censored at 24 months (HR,

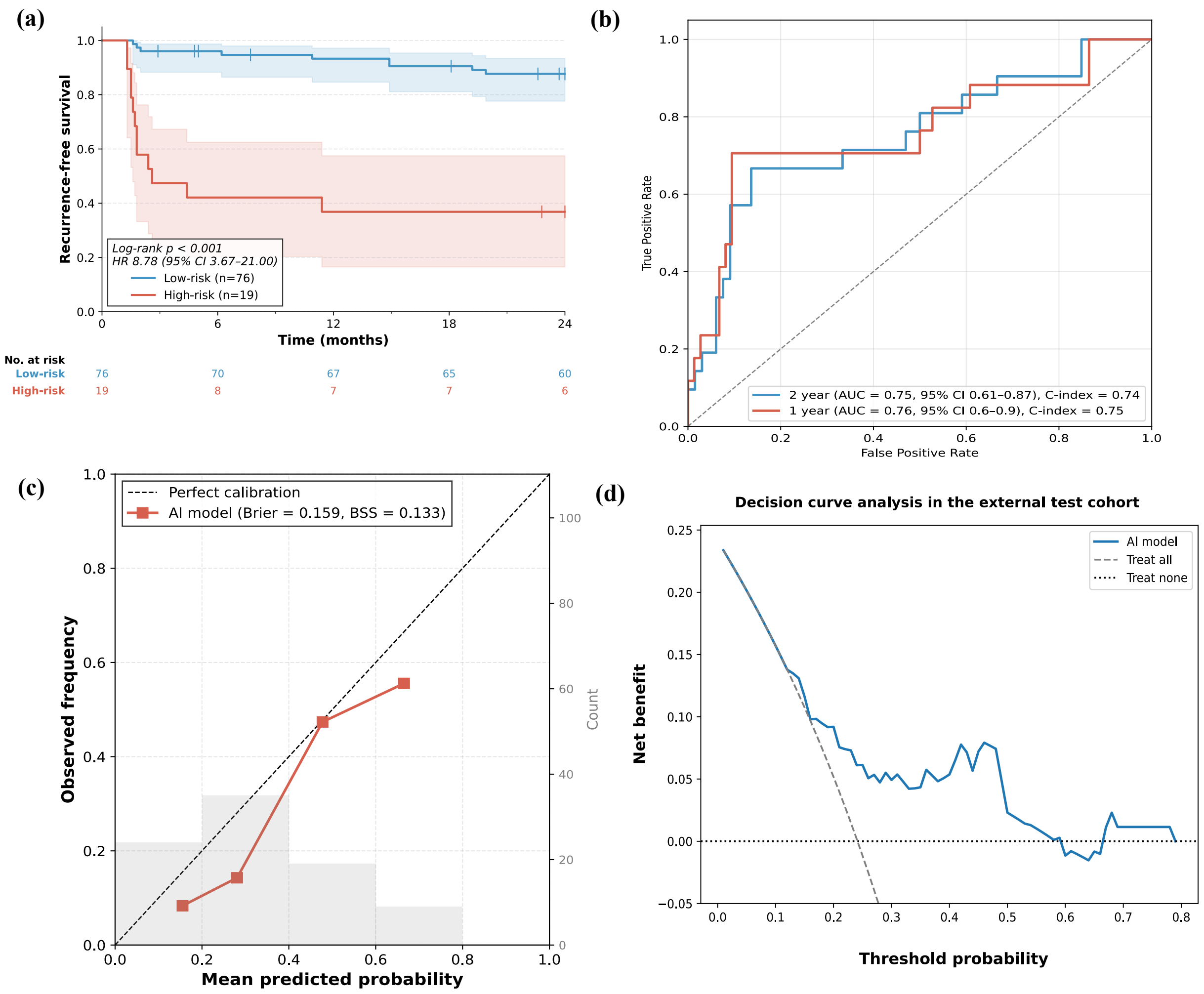

Fig. 5. External validation of the AI model on the CHIMERA cohort (n = 95). (a) Kaplan-Meier recurrence-free survival stratified by the median AI score from the internal development cohort (threshold=0.458), dichotomized into high-risk (n=19) and low-risk (n=76) groups. The high-risk group showed markedly worse survival (≈20% vs. ≈80%) at 24 months. (b) ROC curves for 1-year and 2-year BCR prediction. The AI model achieved AUCs of 0.76 (1-year) and 0.75 (2-year) (c) Calibration plot for 2-year BCR (dashed diagonal = perfect calibration). The AI model remained reasonably calibrated (Brier = 0.159, BSS=0.133), with the gray histogram showing the distribution of predicted probabilities. (d) Decision curve analysis for 2-year BCR; the AI model provided net benefit over the treat-all and treat-none strategies across threshold probabilities of 13%–59%.

8.78; 95% CI, 3.67–21.00).

Figure 5b shows the ROC analysis for the 1- and 2-year endpoints, and the model achieved an AUC of 0.76, and an AUC of 0.75, respectively. These values represent a modest reduction relative to the internal test cohort (e.g., 2-year AUC 0.75 vs. 0.84), likely reflecting that the CHIMERA dataset provides only partial tissue sections per patient rather than the multi-section input for which the model was designed (see Discussion).

The calibration plot showed generally close agreement between predicted and observed BCR risks, with a Brier score of 0.159 and a BSS of 0.133 (Fig. 5c). Decision curve analysis demonstrated a higher net benefit for the AI model than for the treat-all, and treat-none strategies across threshold probabilities of **13%–59%** (Fig. 5d). Collectively, these findings provide preliminary supportive evidence for the transferability of the AI risk score to an independent cohort.

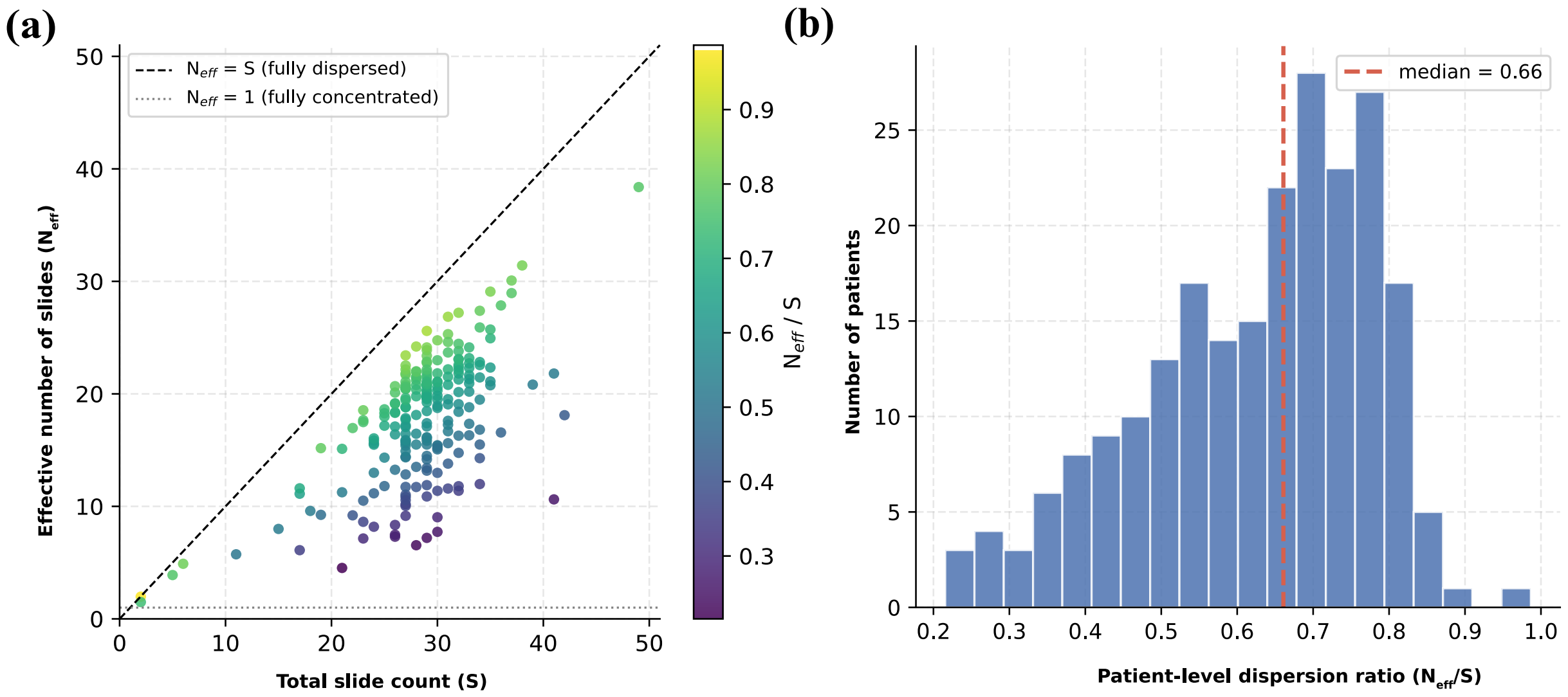

Fig. 6. Dispersion of prognostic attention across whole-gland sections in the internal test cohort (n = 226). (a) $N_{\text{eff}}$ versus the total number of sections ($S$), with each point representing one patient, colored according to the $N_{\text{eff}}/S$ ratio. The dashed line ($N_{\text{eff}} = S$) denotes uniformly dispersed attention across all sections, whereas the dotted line ($N_{\text{eff}} = 1$) represents attention concentrated on a single section. (b) Distribution of the patient-level $N_{\text{eff}}/S$ ratio (median = 0.66, dashed line), suggesting that the effective number of contributing sections corresponded to approximately two-thirds of the available sections.

### *3.6. Assessment of Whole-Gland Attention Dispersion and Pathological Correlates*

To assess whether the model's prognostic signal was concentrated in a small number of sections or distributed across the whole gland, we quantified the slide-level dispersion of the top 10% most-attended patches in the internal test cohort. As shown in Figure 6, most patients exhibited $N_{\text{eff}}$ values well above 1 but below $S$ (Fig. 6a), indicating that prognostic attention was typically distributed across multiple sections, although not uniformly across all available sections. The median effective number of slides ($N_{\text{eff}}$) was 18.8 (IQR, 14.5–21.4), compared with a median slide count of 29 per patient, and the median $N_{\text{eff}}/S$ ratio was 0.66 (Fig. 6b). These findings suggest that the model generally integrates information across multiple sections rather than relying on a single dominant slide for BCR prediction.

In Fig. 7 and Supplementary Fig. S1, two representative cases reviewed by a board-certified pathologist are shown. In the case shown in Fig. 7, all six highest-attention regions were confirmed as carcinoma and exhibited heterogeneous morphologies, including Gleason patterns 3 and 4. Notably, among regions exhibiting Gleason pattern 4, both fused and cribriform glandular architecture were observed. Additional histological features, including perineural invasion and lymphocytic infiltration, were also identified in individual tumor regions. Similar findings were observed for the second case, in which the high-attention regions were likewise confirmed as carcinoma (Supplementary Fig. S1). These findings indicate that the model's spatially distributed attention localizes to multiple, morphologically distinct malignant regions, consistent with the multifocal and heterogeneous nature of prostate cancer.

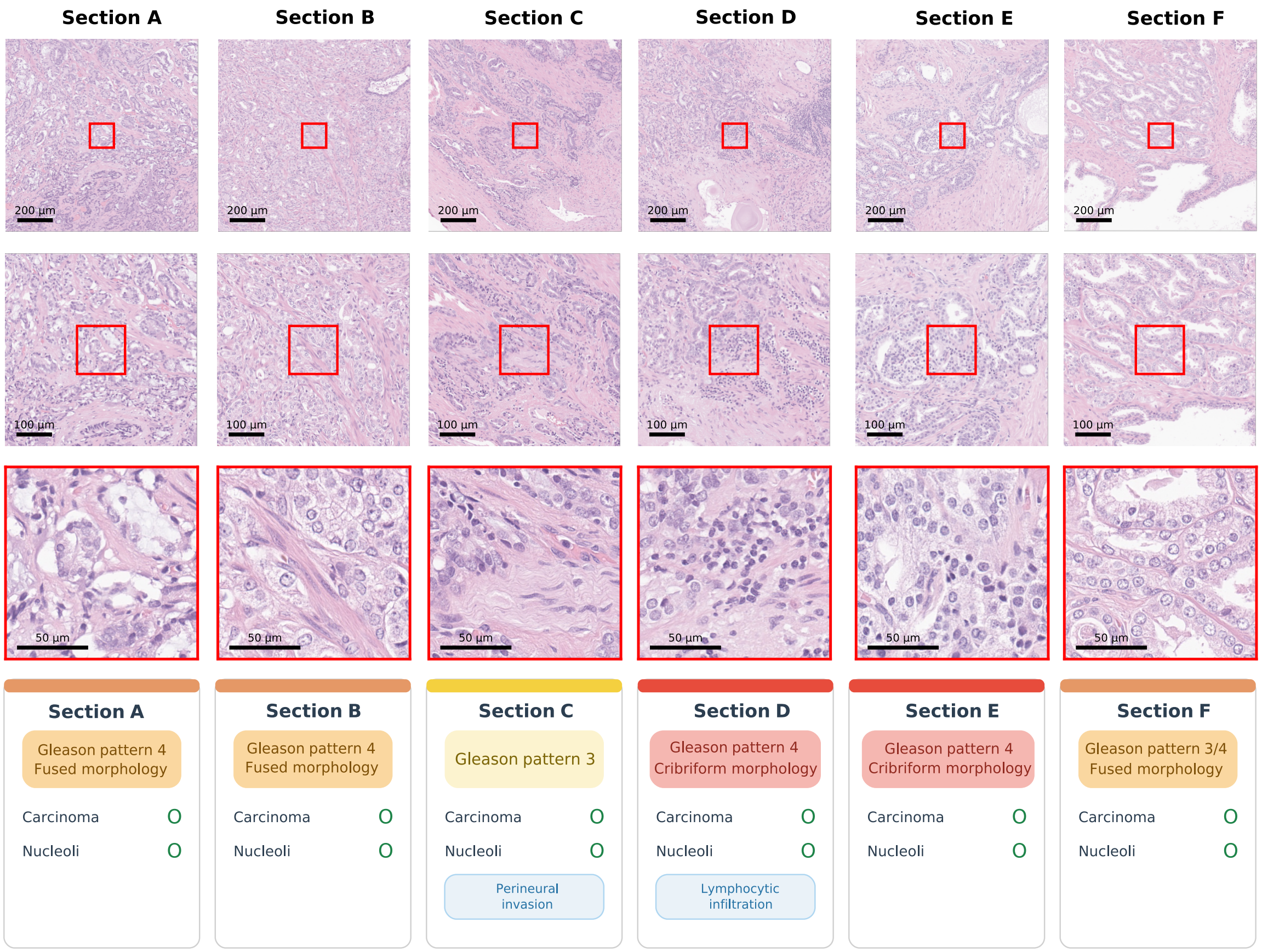

Fig. 7. Pathological review of the highest-attention regions from the six highest-attention sections for a representative patient. For each section, the rows show progressively higher magnification—context fields at ~200 µm and ~100 µm and the peak-attention patch at ~50 µm (scale bars shown; red boxes mark the zoomed region)—and the card below summarizes the board-certified pathologist's assessment. Gleason pattern 4 architectures are annotated as exhibiting fused or cribriform morphology. All six regions were confirmed as carcinoma and exhibited heterogeneous morphologies, with perineural invasion and lymphocytic infiltration noted in individual regions, illustrating that the model's high-attention regions localize to morphologically distinct tumor foci across multiple whole-gland sections.

## 4. Discussion

In this study, we developed a computationally efficient multi-section WSI analysis framework for capturing prostate cancer multifocality across RP specimens to predict BCR within 2 years. The AI-derived risk score was an independent prognostic factor among the covariates examined in multivariable Cox regression analysis (Fig. 3) and showed higher discrimination than established clinical benchmarks (Fig. 4).

The clinical significance of a 2-year recurrence endpoint is well-established in the literature. For instance, Freedland et al. [12] demonstrated that early biochemical recurrence following surgery is associated with adverse oncologic trajectories, indicating that the time to recurrence serves as a proxy for underlying disease aggressiveness. Corroborating studies [36–38] further establish that rapid biochemical progression is indicative of an aggressive tumor biology and correlates with poor downstream prognoses. Consequently, the accurate identification of patients at risk for recurrence within this 2-year window is critical for optimizing postoperative surveillance protocols and facilitating timely adjuvant therapies.

A central premise of this work is that comprehensive spatial coverage of the prostate gland provides prognostic information beyond what single-region analyses can capture. The consistent performance improvement from a single slide (AUC = 0.64) to 19 or more (AUC = 0.81), as shown in Fig. 2d, provides empirical support. Furthermore, the attention-dispersion analysis showed that the model's decisive attention was dispersed across many sections rather than a single dominant region (median $N_{eff}/S$ = 0.66; see Fig. 6), and a board-certified pathologist confirmed that these distributed high-attention regions were localized to multiple, morphologically distinct malignant foci (Fig. 7). Together, these findings align with the well-established tumor heterogeneity across prostate glands [8, 26] and indicate that multifocal features carry complementary prognostic signals lost when only a single section is examined.

The clinical relevance of accurate early BCR prediction is substantial. Current guidelines [3,11] recommend discussing salvage radiotherapy with patients exhibiting adverse pathological features, as timely intervention improves outcomes [9,10]. However, not all such patients will experience BCR, and unnecessary treatment exposes patients to avoidable morbidity. The AI-derived risk score showed separation between the low- and high-risk groups, with 2-year recurrence-free survival rates of 88% and 37%, respectively (Fig. 3b). Although these findings are preliminary, the risk score may have the potential, following prospective validation, to support clinical risk stratification and to inform decisions regarding surveillance intensity or consideration of adjuvant treatment. Further evaluation in larger, prospective cohorts will be required before any clinical application can be established.

Computational efficiency is a prerequisite for clinical translation. Processing whole multi-section slides with the largest inference patch sampling ratio (30%) already required approximately 58 minutes per patient (Fig. 2c). Our patch and slide sub-sampling strategy substantially mitigates this: in the development cohort, reducing the patch sub-sampling ratio to 1% preserved predictive performance (AUC = 0.82) at 9.4 minutes per patient, and further slide sub-sampling achieved comparable performance (AUC = 0.81) with as few as 19 slides in approximately 6.3 minutes (Fig. 2d). Ultimately, this streamlined processing time suggests the potential feasibility of integrating our model into routine clinical workflows.

External validation using the CHIMERA dataset provided preliminary evidence that the AI-derived risk score may be transferable to an independent cohort; however, these findings should be interpreted cautiously. A notable limitation of CHIMERA was the sparse representation of the prostate gland, with an average of only 1.9 slides per patient. This limited tissue coverage may have been insufficient to capture the spatially heterogeneous nature of prostate cancer and contributed to the reduced predictive performance observed in external validation (Fig. 5b). Additional differences between the internal and external test cohorts included the use of different slide scanners (Aperio GT 450 DX versus 3DHISTECH) and different definitions of biochemical recurrence (PSA ≥ 0.2 ng/mL versus PSA ≥ 0.1 ng/mL). Therefore, the validation results should be regarded as preliminary supportive evidence rather than definitive validation of the model's generalizability, particularly for the full multi-section framework.

Several limitations should be acknowledged. First, the internal cohort was collected from a single institution; multi-institutional validation on larger, multi-section cohorts, reporting center, scanner, and staining heterogeneity, is needed to establish generalizability. Second, the framework employs a standard attention-based MIL [28] with the UNI [34] foundation model without considering inter-slide spatial relationships. Future work could explore hierarchical attention mechanisms to address this [39–42]. In addition, incorporating prostate cancer-specific foundation models may further improve performance and robustness [43]. Third, comparison with prior methods was limited to a few clinical models. This is because existing pathology-based AI studies [19,20,21,22] have not, to our knowledge, released their model weights for benchmarking, and notably, there is a lack of prior models designed to utilize multi-section whole slide images. Finally, the present study focused on early BCR at 1- and 2-year time points. Extending the framework to longer-term recurrence horizons (e.g., 5–10 years) using a time-to-event formulation, such as a Cox-based loss [44] or DeepHit [45], represents an important direction for future work.

## 5. Conclusion

We presented a computationally efficient multi-section WSI framework for capturing prostate cancer multifocality and predicting early BCR. Through patch- and slide-level sub-sampling, inference with only 1% of patches and as few as 19 slides preserved predictive accuracy while reducing per-patient processing time from approximately 58 to 6 minutes. Under this configuration, the model achieved strong discrimination for both 1- and 2-year BCR in the internal test cohort (AUC = 0.84), with discrimination, calibration, and clinical net benefit comparable to or exceeding those of established clinical benchmarks; in multivariable Cox analysis, the AI-derived risk score was an independent prognostic factor among the covariates examined (HR = 3.27, C-index = 0.79). The attention-dispersion analysis of AI showed that prognostic attention was distributed across multiple whole-gland sections and supported its multifocality-awareness. External validation provided preliminary evidence of transferability. Overall, the framework combines computational efficiency with strong prognostic performance, offering a promising approach to support post-operative management in prostate cancer, pending prospective, multi-institutional validation.

## Acknowledgements

Approval for the use of the study data was granted by the Institutional Review Board of Pusan National University Hospital (IRB No. 2201-006-110, date: 2022/01/06). Patient identifiers were permanently removed to ensure that all subsequent analyses were conducted anonymously.

### Funding

This work was supported by the National Research Foundation of Korea (NRF) grant funded by the Korea government (MSIT) (No. NRF-2023R1A2C1003760) and by the Bio & Medical Technology Development Program of the National Research Foundation (NRF) funded by the Korean government (MSIT) (No. RS-2023-00223277).

## Author Contributions

Y.C. and D.S. contributed equally to this work. Y.C. and D.S. conceived and designed the study, developed the AI analysis workflow, and conducted all experiments. S.H., G.L., and S.P. assisted with data processing and implementation. W.Y.P., a board-certified pathologist, performed independent pathological review of the AI-identified key regions, confirming their malignancy and characterizing their morphologically diverse multifocal features. B.P. provided clinical consultation during this study. J.L. and H.K.H. provided clinical data, pathological expertise, and supervised clinical aspects of the study. J.P. and H.K.H. supervised the overall study. All authors reviewed and approved the final manuscript.

## Competing Interests

D.S. and J.P. are officers of OmixAI Co., Ltd. and Oncocross Co., Ltd. Y.C., S.H., and G.L. are employees of OmixAI Co., Ltd. S.P. is an employee of Oncocross Co., Ltd. The remaining authors declare no competing interests.

## Data Availability

The internal dataset used in this study contains sensitive patient information and cannot be publicly shared due to confidentiality restrictions imposed by the Pusan National University Hospital Institutional Review Board. Access to de-identified data may be considered upon reasonable request and following ethics committee approval. The CHIMERA external validation dataset is publicly available at https://chimera.grand-challenge.org/.

## Code Availability

The code used in this study is publicly available on GitHub (https://github.com/Radisen-Panthera/Recurrence_Prediction_MIL_PNU_Official_code) and archived on Zenodo (https://doi.org/10.5281/zenodo.21865904). The trained model weights and the underlying data are available from the corresponding author upon reasonable request.

## Ethics Declarations

### *Approval for Human Experiments*

This study was approved by the Institutional Review Board of Pusan National University Hospital (IRB No. 2201-006-110, date: 2022/01/06) and was conducted in accordance with the Declaration of Helsinki and its later amendments.

### *Consent to Participate*

The requirement for informed consent was waived by the Institutional Review Board of Pusan National University Hospital due to the retrospective nature of the study. Patient identifiers were permanently removed to ensure anonymity prior to analysis.